\pdfoutput=1

\documentclass[11pt]{article}

\usepackage[]{emnlp2021}

\usepackage{times}
\usepackage{latexsym}

\usepackage[T1]{fontenc}

\usepackage[utf8]{inputenc}

\usepackage{microtype}
\usepackage{booktabs}
\usepackage{graphicx}
\usepackage{amsmath}
\usepackage{multirow} 
\usepackage{enumitem}
\usepackage{url}
\usepackage{subfigure}
\usepackage{tabularx}

\title{Enhancing Visual Dialog Questioner with Entity-based Strategy Learning and Augmented Guesser}

\author{Duo Zheng\textsuperscript{1}\thanks{ \ \ Equal contribution. Work was done when Zheng and Xu were interning at Pattern Recognition Center, WeChat AI, Tencent Inc, China.}, Zipeng Xu\textsuperscript{1}\footnotemark[1], Fandong Meng\textsuperscript{2}, Xiaojie Wang\textsuperscript{1}\thanks{ \ \ Xiaojie Wang is the corresponding author.},\\
\bf {Jiaan Wang\textsuperscript{3}, Jie Zhou\textsuperscript{2}} \\
\textsuperscript{1}Beijing University of Posts and Telecommunications \\
\textsuperscript{2}Pattern Recognition Center, WeChat AI, Tencent, China \\
\textsuperscript{3}Soochow University, Suzhou, China \\
\texttt{\{zd, xuzp, xjwang\}@bupt.edu.cn}, \texttt{jawang1@stu.suda.edu.cn}\\
\texttt{\{fandongmeng,withtomzhou\}@tencent.com} \\
}

\begin{document}
\maketitle
\begin{abstract}
Considering the importance of building a good Visual Dialog (VD) Questioner, many researchers study the topic under a Q-Bot-A-Bot image-guessing game setting, where the Questioner needs to raise a series of questions to collect information of an undisclosed image. Despite progress has been made in Supervised Learning (SL) and Reinforcement Learning (RL), issues still exist. Firstly, previous methods do not provide explicit and effective guidance for Questioner to generate visually related and informative questions. Secondly, the effect of RL is hampered by an incompetent component, i.e., the Guesser, who makes image predictions based on the generated dialogs and assigns rewards accordingly. To enhance VD Questioner: 1) we propose a \textbf{R}elated \textbf{e}ntity \textbf{e}nhanced \textbf{Q}uestioner (ReeQ) that generates questions under the guidance of related entities and learns entity-based questioning strategy from human dialogs; 2) we propose an \textbf{Aug}mented \textbf{G}uesser (AugG) that is strong and is optimized for the VD setting especially. Experimental results on the VisDial v1.0 dataset show that our approach achieves state-of-the-art performance on both image-guessing task and question diversity. Human study further proves that our model generates more visually related, informative and coherent questions. 
\end{abstract}

\section{Introduction}

Visual Dialog (VD), which expects AI agents to conduct visually related dialog, has attracted growing interests due to its research significance and application prospects. Most of the work~\cite{DBLP:journals/corr/LuKYPB17,2018Recursive,2019Multi,2020DMRM,agarwal-etal-2020-history,2020Efficient,2021GOG} pays attention to modeling an Answerer agent. However, it is also important to model a VD Questioner agent that can constantly ask visually related and informative questions.

Previous researches~\cite{Das2017LearningCV,Murahari2019ImprovingGV,2019Building} have explored building open-domain VD Questioner under a Q-Bot-A-Bot image-guessing game setting, namely GuessWhich~\cite{Das2017LearningCV}. Given an undisclosed image, GuessWhich can be regarded to have two stages: 1) Dialog generation stage: Q-Bot (Questioner, who only knows a caption of the image at first) successively asks questions to collect information about the image, and A-Bot (Answerer, who can see the image) answers the questions. 2) Guess stage: Q-Bot guesses the target image based on the generated dialog. Corresponding to the two stages, Q-Bot has two roles, i.e, Question Generator (QGen) and Guesser\footnote{We borrow the two concepts from GuessWhat?!~\cite{2016GuessWhat} to clarify the two correspondingly identical roles of Q-Bot in the GuessWhich setting.}. Besides Supervised Learning (SL), previous methods ~\cite{Das2017LearningCV,Murahari2019ImprovingGV,2019Building} introduce Reinforcement Learning (RL) to further fine-tune the agent. Though progress has been made, issues still exist.

Firstly, previous work does not provide explicit and effective guidance to generate visually related and informative questions. To encourage diverse questions,~\citet{Murahari2019ImprovingGV} penalize the similarity in successive textual dialog hidden states. But this method can not promise the diverse questions are visually related. To ask visually related questions,~\citet{2019Building} retrieve the most likely image at each round to provide Questioner with visual information. Yet, an image contains many contents while the method does not provide explicit guidance for Questioner to ask about which one.

Secondly, the reward in RL is not efficient due to an incompetent Guesser, hampering the effect of RL optimization. At each round of the dialog, Guesser makes an image feature prediction based on current dialog, then the reward is assigned to encourage the reduce of the distance between image feature prediction and target image feature. The efficiency of reward relies on the performance of Guesser heavily. However, previous Guessers' performance is limited. This results from a cooperative training setting, where Guesser shares the same encoder with QGen and is optimized jointly. As illustrated in Tab.~\ref{ratio}, using previous method, it is impossible to simultaneously obtain a good QGen and a good Guesser. Conventionally, since QGen's performance is of higher priority to be concerned, the performance of Guesser consequently becomes inferior. As they use this limited Guesser to assign reward in RL, the reward is likely to be uncertain and thus inefficient. The effect of RL optimization is hampered consequently. Further progress requires a stronger Guesser to assign reliable rewards.

\begin{table}[]
\centering
\small
\setlength{\tabcolsep}{2.5mm}{
\begin{tabular}{lcc} 
\toprule
$\alpha$ & PMR$\uparrow$ &Unique questions$\uparrow$  \\ \hline
500 & 93.91 & \textbf{6.39} \\
1000 & 96.22 & 6.22 \\
2000 & 96.48 & 6.09 \\
4000 & \textbf{96.60} & 4.36 \\
\bottomrule
\end{tabular}}
\caption{In previous method~\cite{Das2017LearningCV}, a good QGen with higher unique questions, is together with a limited Guesser with lower PMR (Percentile Mean Rank). $\alpha$ is the loss ratio of Guesser to QGen in cooperative training.\protect\footnotemark}
\label{ratio}
\end{table}
\footnotetext{As they find change of the loss ratio will lead to different results on PMR, we conduct experiments that train the model with different loss ratios using their code~\cite{modhe2018visdialrlpytorch}.}

To remedy above issues, we propose a \textbf{R}elated \textbf{e}ntity \textbf{e}nhanced \textbf{Q}uestioner (ReeQ) and an \textbf{Aug}mented \textbf{G}uesser (AugG) to enhance the VD Questioner in both SL and RL. 
ReeQ is a Questioner that explicitly uses related entities as guidance to generate questions and learns entity-based questioning strategy through large-scale human dialogs. In concrete, ReeQ firstly uses the image caption to retrieve related entities, which are preprocessed to be related to the target image; then at each round of the dialog, it selects which entity to ask about according to current dialog condition; lastly, it uses the selected entity as a hint to guide question generation.
The related entities help ReeQ to ask visually-related questions while questioning strategy-learning enables it to ask constantly informative questions.
AugG is a strong Guesser that is optimized with a special consideration for the VD setting. 
Specifically, we separately train the AugG with a hinge loss that incorporates hard negative samples.
In particular, we introduce the competitive VD-oriented negative samples, which are images that contain alike visual contents related to the caption of target image, so as to enforce more distinguishable image feature predictions from the model, especially when dialog contexts are similar.
In RL, we use AugG to assign reliable rewards to further improve the Questioner.

We evaluate our method on the VisDial v1.0 dataset~\cite{Das2017VisualD}. Experimental results show that our approach achieves state-of-the-art (SOTA) performance on both the image-guessing task and question diversity. Human study indicates that our Questioner generates more visually related, informative and coherent questions as compared to previous strong baselines.

Our main contributions\footnote{We release our code on \url{https://github.com/zd11024/Entity_Questioner}.} are concluded as follows:
\begin{itemize}
\setlength{\itemsep}{3pt}
\setlength{\parsep}{0pt}
\setlength{\parskip}{0pt}
\item We propose a \textbf{R}elated \textbf{e}ntity \textbf{e}nhanced \textbf{Q}uestioner (ReeQ) for Visual Dialog. ReeQ generates questions using related entities as guidance and learns entity-based questioning strategy from human dialogs.
\item We propose an \textbf{Aug}mented \textbf{G}uesser (AugG) and use it to serve as an efficient component in RL to assign reliable rewards.
\item We conduct experiments on the VisDial v1.0 dataset and achieve SOTA performance on both the image-guessing task and question diversity. Our Questioner outperforms previous methods on multiple criteria.
\end{itemize}

\section{Background}
GuessWhich~\cite{Das2017LearningCV} is an interactive Q-Bot-A-Bot image-guessing task. Q-Bot, who only knows the caption of an undisclosed image $I$ at first, needs to ask a series of questions and guess the target image. A-Bot, who can see the image, answers accordingly.
In this section, we formally introduce the modeling of Q-Bot and A-Bot in previous methods~\cite{Das2017LearningCV, Murahari2019ImprovingGV}, as well as the training paradigm.
\subsection{Model} \label{background_model}
\paragraph{Q-Bot.} At round $t$, Q-Bot generates the question $q_{t+1}$ and makes an image feature prediction $\mathbf{\hat y_t}$ based on the dialog history $H_{t}=\{c,(q_1,a_1), \dots,(q_{t},a_{t})\}$, where $c$ is the caption of the target image. It consists of Context Encoder, Feature Regression Network, and Question Decoder. After the dialog history $H_{t}$ is encoded into a dense vector, the Feature Regression Network is used to make an image feature prediction, while the Question Decoder is used to generate question $q_{t+1}$.

\begin{itemize}[leftmargin=*]
\setlength{\itemsep}{0pt}
\setlength{\parsep}{0pt}
\setlength{\parskip}{0pt}
\item \textbf{Context Encoder}: The Context Encoder consists of fact encoder and history encoder, both are two-layer LSTM~\cite{1997Long}. At round $t$, fact encoder encodes the question-answer pair $(q_t, a_t)$ into the fact representation $\mathbf{f_t}$, then history encoder encodes $\mathbf{f_t}$ into the history representation $\mathbf{h_t}$.
\item \textbf{Feature Regression Network}: An MLP that uses history representation $\mathbf{h_t}$ to make the image feature prediction $\mathbf{\hat y_t}$ at round $t$.
\item \textbf{Question Decoder}: A two-layer LSTM that decodes the question $q_{t+1}$ given the history representation $\mathbf{h_{t}}$.
\end{itemize}

Corresponding to the two roles of Q-Bot, Context Encoder and Question Decoder form the Question Generator (QGen), while Context Encoder and Feature Regression Network form the Guesser.

\paragraph{A-Bot.} Given image $I$, dialog history $H_{t}$ and question $q_{t+1}$, A-Bot generates the answer $a_{t+1}$. A-Bot consists of a multi-modal context encoder and a decoder.

\subsection{Training} \label{background_training}
Previous methods use a two-stage training paradigm: Q-Bot and A-Bot are firstly pre-trained through Supervised Learning (SL), then fine-tuned through Reinforcement Learning (RL).


\paragraph{SL.} Q-Bot and A-Bot are respectively optimized in SL. Q-Bot (QGen and Guesser) is optimized with multi-task loss: a Cross-Entropy (CE) loss $\mathcal{L}_{CE}=\sum_{t} log(p(q_{t+1}|\mathbf{h_t}))$ to optimize QGen and a Mean Square Error (MSE) loss $\mathcal{L}_{MSE}=\sum_{t} \Vert \mathbf{\mathbf{y^{gt}-\hat y_t}} \Vert_2^2$, where $\mathbf{y^{gt}}$ is the image feature of $I$, to optimize Guesser. A-Bot is optimized with a similar CE loss.

\begin{figure*} \centering    
\subfigure[ReeQ] {
 \label{fig:a}     
\includegraphics[width=1.2\columnwidth]{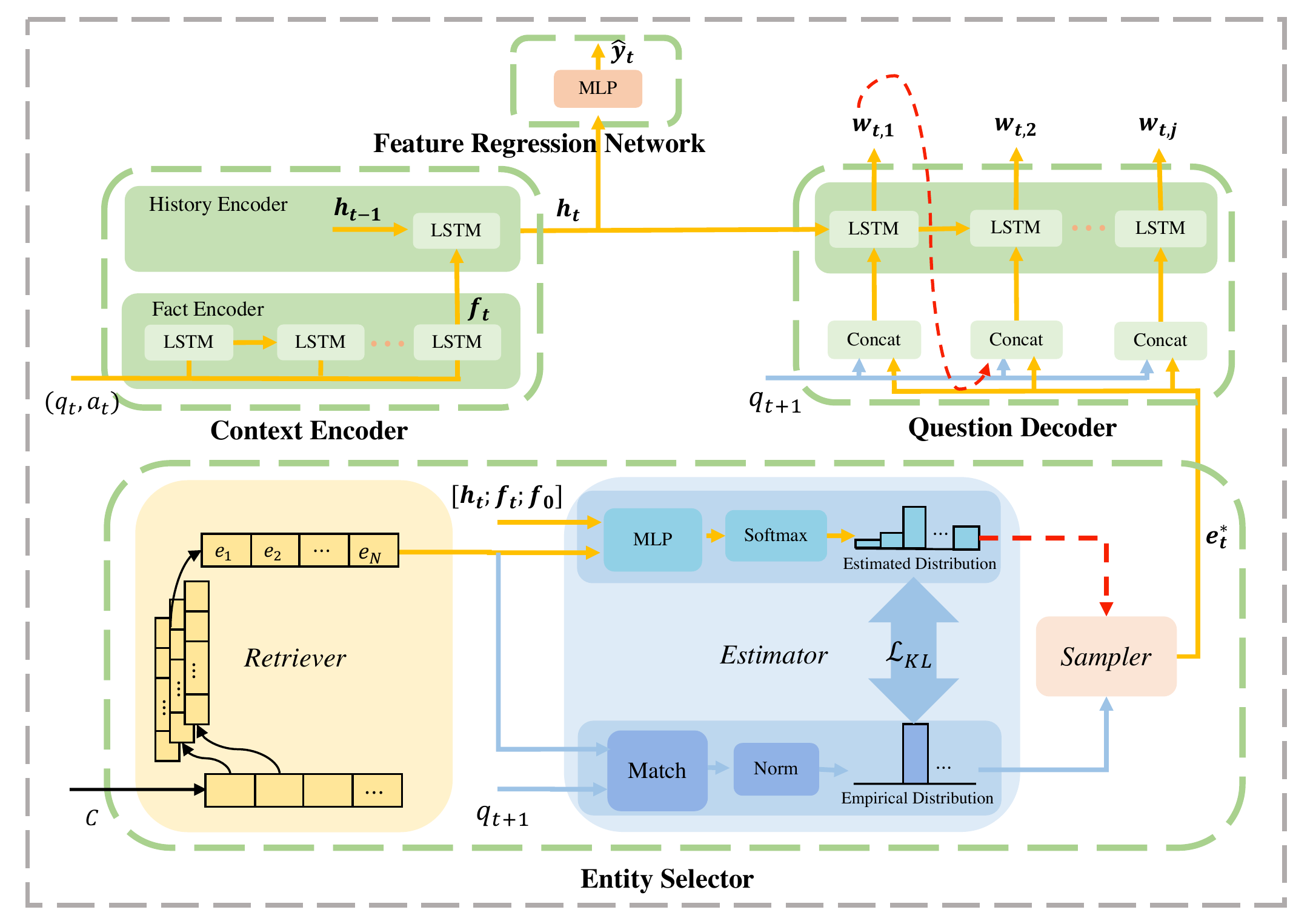}  
}     
\subfigure[AugG] { 
\label{fig:b}     
\includegraphics[width=0.6\columnwidth]{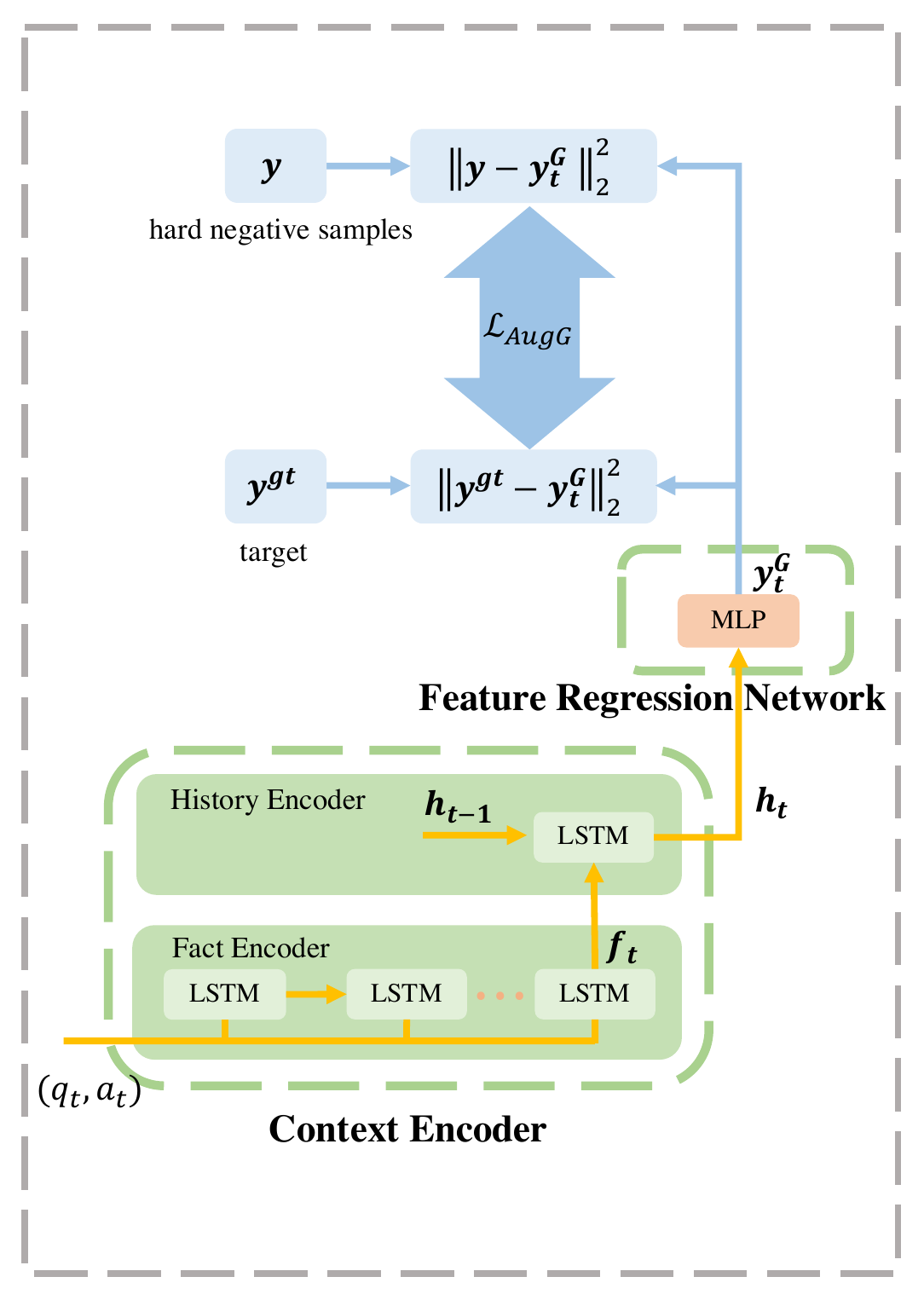}     
}    
\caption{Illustration of our model architecture. (a) ReeQ. ReeQ contains four modules: Context Encoder, Entity Selector, Question Decoder and Feature Regression Network. Orange and blue line indicate calculation path in training. Orange and red line indicate the calculation path in inference. (b) AugG. AugG contains two modules: Context Encoder and Feature Regression Network, and is augmented with hard negative samples.}
\label{model_architecture}
\end{figure*}

\paragraph{RL.} Q-Bot and A-Bot are jointly optimized in RL. Q-Bot and A-Bot interact with each other and are awarded by reward $r_t=\Vert \mathbf{y^{gt}} - \mathbf{\hat y_{t}} \Vert_2^2 - \Vert \mathbf{y^{gt}} - \mathbf{\hat y_{t+1}} \Vert_2^2$. Given the Q-Bot state $S_{t}^Q$ and A-Bot state $S_{t}^A$, dialog policies for Q-Bot and A-Bot are formulated as $\pi_{\theta_Q}(q_{t} |S_{t}^Q)$ and $\pi_{\theta_A}(a_{t}|S_{t}^A)$, respectively. The action of Q-Bot and A-Bot is to select next token from the vocabulary $\mathcal{V}$. REINFORCE\cite{1992Simple} algorithm is applied to update agents' parameters with the policy gradients formulated as $E_{\pi_Q,\pi_A}r_t \nabla_\theta{_Q} log(\pi_Q(q_t|S_t^Q))$ and $E_{\pi_Q,\pi_A}r_t \nabla_\theta{_A} log(\pi_A(a_t|S_t^A))$.


To conclude: 1) previous QGen follows a sequence-to-sequence fashion to generate questions and lacks a clear questioning strategy; and 2) reward in RL relies on a limited Guesser, that has been compromised in the eclectic training result of optimizing QGen and Guesser cooperatively.

\section{Approach}
In this section, we introduce the \textbf{R}elated \textbf{e}ntity \textbf{e}nhanced \textbf{Q}uestioner (ReeQ), \textbf{Aug}mented \textbf{G}uesser (AugG) and training approach. ReeQ generates questions under the guidance of related entities and learns entity-based questioning strategy from human dialogs. AugG is a strong guesser and assigns rewards during the RL optimization process.

\subsection{Related Entity Enhanced Questioner}\label{ReeQ}
As illustrated in Fig.~\ref{model_architecture} (a), ReeQ consists of four modules: Context Encoder, Entity Selector, Question Decoder and Feature Regression Network.
To generate a question at round $t$: firstly, Context Encoder encodes dialog history $H_t$ into a history representation $\mathbf{h_t}$; then, Entity Selector selects a specific entity $e_t^*$ to ask about at this round; lastly, Question Decoder generates the question $q_{t+1}$ with the selected entity $e_t^*$ as guidance.

Context Encoder and Feature Regression Network are the same as previous work (see Sec.~\ref{background_model}). We introduce the Entity Selector and the Question Decoder in Sec.~\ref{entity_selctor} and \ref{question_decoder}, respectively.  

\subsubsection{Entity Selector}
As illustrated in Fig.~\ref{model_architecture} (a), Entity Selector contains three components, i.e., \textit{Retriever}, \textit{Estimator} and \textit{Sampler}.
~\label{entity_selctor}
Initially, \textit{Retriever} retrieves a series of candidate entities using the image caption. At each round of the dialog, \textit{Estimator} estimates a probability distribution on candidate entities w.r.t. the probable entities to ask about. Lastly, \textit{Sampler} samples an entity based on the estimated distribution.

\paragraph{Retriever.}\label{retriever}
\textit{Retriever} uses image caption to retrieve the related entities in advance. As a prerequisite, we build entities-to-entities indexes from the entities in captions to the entities in dialogs, based on the co-occurrences in training data. While in use, for each dialog instance, we firstly extract the entities in caption, then use them as queries to retrieve a list of candidate entities, i.e., $E=\{e_1,e_2,\cdots, e_N\}$, from the established indexes. To assure the relevancy, we retain the top $N$ entities with the highest co-occurrence frequency. More details are given in Appendix~\ref{entity_retrieval}.

\paragraph{Estimator.}
\textit{Estimator} estimates a probability distribution on candidate entities, w.r.t. the probable entities to ask about at each round of the dialog.

The estimated distribution $p^{est}_t$ is derived conditioning on current dialog, concretely the history representation $\mathbf{h_{t}}$, fact representation $\mathbf{f_{t}}$ and caption representation $\mathbf{f_0}$. We formulate this step as:
\begin{gather}
\setlength{\abovedisplayskip}{5pt}
\setlength{\belowdisplayskip}{10pt}
v_i=tanh([\mathbf{h_{t}}; \mathbf{f_{t}}; \mathbf{f_0}]\mathbf{W^Q}+\mathbf{e_i}\mathbf{W^K})\mathbf{W^A}, \\
p^{est}_t(e_i)=\mathop{Softmax}(v_i),
\end{gather}
where $\mathbf{W^Q}, \mathbf{W^K}$ and $\mathbf{W^A}$ are learnable parameters; $\mathbf{e_i}$ is entity representation encoded by a LSTM as an entity may include more than one word.

To learn the distribution, we establish empirical distribution $p^{emp}_t$ from human dialog in the training data and propose an objective to encourage estimated distribution to approximate empirical distribution.
In specific, empirical distribution is obtained by matching golden question $q_{t+1}$ and candidate entities as follows:
\begin{gather}
\setlength{\abovedisplayskip}{3pt}
\setlength{\belowdisplayskip}{3pt}
p^{emp}_t(e_i)=\mathop{Norm}(Match(e_i,q_{t+1})),
\end{gather}
where $Match(e_i, q_{t+1})$ is 1 when $e_i$ could match a sub-string of $q_{t+1}$, otherwise 0; $Norm(\cdot)$ is a sum-normalization to normalize the matching result as probability distribution.

Further, we minimize the KL divergence between empirical distribution and estimated distribution throughout the dialog, so as to learn the questioning strategy from human dialog. The KL loss is formulated as:
\begin{equation}
\setlength{\abovedisplayskip}{3pt}
\setlength{\belowdisplayskip}{3pt}
\mathcal{L}_{KL}=\sum_{t,i} D_{KL}(p_t^{emp}(e_i)||p_t^{est}(e_i)).
\label{klLoss}
\end{equation}

Eq.~\ref{klLoss} is optimized during training. While in inference, \textit{estimator} only needs to calculate the estimated distribution.

\paragraph{Sampler.}
\textit{Sampler} samples an entity based on the distribution given by \textit{Estimator} -- empirical distribution during training while estimated distribution during inference. We formulate this step as:
\begin{gather}
\setlength{\abovedisplayskip}{3pt}
\setlength{\belowdisplayskip}{3pt}
sample \ e_{t}^* \sim \left\{
\begin{aligned}
&~~p^{emp}_t,  & \quad if \ training, \\
&~~p^{est}_t,  & \quad if \ inference.
\end{aligned}
\right.
\end{gather}

To further refine the questioning strategy during inference, we propose a \textit{limit-sampling rule} which limits the sampled times of each entity.
In concrete, we count the sampled times $c_{t}^i$ of each entity ($c_0^i\!=\!0$), and when $c_{t}^i$ reaches the upper bound $B$, the corresponding entity will be masked. Accordingly, the refined estimated distribution is:
\begin{equation}
\setlength{\abovedisplayskip}{3pt}
\setlength{\belowdisplayskip}{3pt}
p^{est}_t(e_i)\!=\!\mathop{MaskedSoftmax}(I[c_{t}^i <\!B] v_i).
\label{upb}
\end{equation}
The sampled times is updated as follows:
\begin{equation}
\setlength{\abovedisplayskip}{3pt}
\setlength{\belowdisplayskip}{3pt}
c_{t+1}^i=c_{t}^i+I[e_{t}^*=e_i].
\label{upb}
\end{equation}
where $I[\cdot]$ equals 1 when the expression in square brackets is true, else 0.

\subsubsection{Question Decoder}\label{question_decoder}
Question Decoder is a two-layer LSTM that generates next question using the selected entity as a hint. At each time step $j$, we concatenate the previously generated word embedding $\mathbf{w_{t,j-1}}$ with the selected entity representation $\mathbf{e_t^*}$ as input.

With $\mathbf{h_{t, j}^D}$ ($\mathbf{h_{t,0}^D=h_t}$) denoting the hidden states of the decoder at the time step $j$, we formulate the decoding step as:
\begin{gather}
\label{decoder_eq_1}
\setlength{\abovedisplayskip}{3pt}
\setlength{\belowdisplayskip}{3pt}
\mathbf{h_{t,j}^{D}}=LSTM^D([\mathbf{w_{t,j-1}};\mathbf{e_{t}^{*}}], \mathbf{h_{t,j-1}^D)}, \\
p(w_{t,j}|\mathbf{h_{t,j}^{D}})=\mathop{softmax}(\mathbf{h_{t,j}^{D}}\mathbf{W^{D}}).
\end{gather}

\subsection{Augmented Guesser} \label{AugG}
We establish the Augmented Guesser (AugG) using the same two modules, i.e., Context Encoder and Feature Regression Network, as shown in Fig.~\ref{model_architecture} (b). At round $t$, given dialog history $H_t$, AugG makes the image feature prediction $\mathbf{y_t^G}$.

To enable a strong AugG, we provide two types of negative samples during training.
The first is the VD-oriented negative samples, which are images retrieved by the distinctive caption in each dialog instance (see details in Appendix~\ref{negative_samples}).
Thus, the VD-oriented negative image samples have alike visual semantics with the target image. Such negative samples enforce more distinguishable image predictions under similar dialog context.
The second is the stochastic negative samples in mini-batch, drawing on the use of negative mining in other tasks \cite{Schroff_2015_CVPR,Wu_2017_ICCV,2017VSE}.
Sec.~\ref{training_AugG} introduces the detailed loss function.

\begin{table*}[]
\renewcommand\arraystretch{0.9}
\centering
\setlength{\tabcolsep}{2.5mm}{
\begin{tabular}{c|c|cc|cccccc} \hline
\#         &        & QGen & Guesser & MRR $\uparrow$ & R@1 $\uparrow$  & R@5 $\uparrow$ &R@10 $\uparrow$ & Mean $\downarrow$  & PMR  $\uparrow$\\ \hline
1 &\multirow{5}{*}{SL}  & DasQ$\dagger$ & DasG$\dagger$ & 7.80 & 2.56  & 9.49  & 17.87   & 127.84  & 93.83  \\
2 &                 & DivQ$\dagger$ & DivG$\dagger$ & 10.73 & 3.39  & 14.82  & 25.29   & 87.92  & 95.73  \\ 
                
3 &                & DasQ$\dagger$ & AugG & 25.65  &  16.30 & 40.50  & 55.43   & 28.57  & 98.76 \\ 
4 &                & DivQ$\dagger$  & AugG & 31.59 & 19.14  & 44.96  & 59.06   & 22.03  & 98.93  \\ 
5 &                & ReeQ  & AugG & 31.21 & 17.78  & 45.01  & 59.98   & 20.60  & 99.00  \\ 
                \hline
6 & \multirow{5}{*}{RL}  & DasQ$\dagger$ & DasG$\dagger$ & 7.54 &2.18  & 9.78  & 17.05   & 125.07  & 93.94  \\
7 &                 & DivQ$\dagger$ & DivG$\dagger$ & 10.79 & 3.39  & 15.69  & 25.33  & 89.28  & 95.67 \\
8 &                 & DasQ$\dagger$ & AugG & 29.52 &  16.57 & 42.68  & 57.99  & 25.36  & 98.77   \\ 
9 &                & DivQ$\dagger$  & AugG& 31.08  & 17.93  & 44.91  & 60.41  & 22.35  & 98.91  \\
10 &                & ReeQ   & AugG & \textbf{33.65} & \textbf{19.91} & \textbf{48.50} & \textbf{62.94}  & \textbf{18.05}  &  \textbf{99.13}  \\ 
                 \hline
\end{tabular}}
\caption{Comparing results on image-guessing task. $\dagger$ represents that the evaluated models are from ~\cite{Murahari2019visdialdivpytorch}. $\uparrow$ indicates higher is better. $\downarrow$ indicates lower is better.}
\label{image-guessing-performance}
\end{table*}

\subsection{Training}
Our training is two-stage: 1) firstly train ReeQ, AugG and A-Bot through Supervised learning (SL); then 2) jointly fine-tune ReeQ and A-Bot through Reinforcement Learning (RL) with the reward assigned by AugG.
~\label{training}
\subsubsection{Supervised Learning}
\paragraph{Training for ReeQ.}
Similar to previous work, ReeQ is optimized with multi-task loss that includes $\mathcal{L}_{CE}$ and $\mathcal{L}_{MSE}$ as in Sec.~\ref{background_training}.
Besides, as given in Eq.~\ref{klLoss}, $\mathcal{L}_{KL}$ is to make the estimated distribution approximate the empirical distribution. Thus, the loss function for ReeQ is:
\begin{equation}
\setlength{\abovedisplayskip}{3pt}
\setlength{\belowdisplayskip}{3pt}
\mathcal{L}_{ReeQ}= \mathcal{L}_{CE}+ \beta \mathcal{L}_{MSE} + \gamma \mathcal{L}_{KL},
\end{equation}
where $\beta$ and $\gamma$ are hyper-parameters.

\paragraph{Training for AugG.}\label{training_AugG}
The loss to optimize AugG is based on $\alpha$-margin max-of-hinges loss \cite{2017VSE} and incorporates two types of negative samples (Sec.~\ref{AugG}). We formulate the loss as:
\begin{equation}
\setlength{\abovedisplayskip}{0pt}
\setlength{\belowdisplayskip}{0pt}
\begin{aligned}
\mathcal{L}_{AugG}&\!=\!\sum_{t}\!\mathop{max}_{y\in Y}[\alpha\!+\!\Vert \!\mathbf{y^{gt}}\!-\!\mathbf{y^{G}_t}\!\Vert_2^2\!-\!\Vert\! \mathbf{y}\!-\!\mathbf{y_t^G}\!\Vert_2^2]_{+}\\
&\!+\!\sum_{t}\!\mathop{max}_{y'\in Y'}[\alpha\!+\!\Vert \!\mathbf{y^{gt}}\!-\!\mathbf{y_t^{G}}\!\Vert_2^2\!-\!\Vert\! \mathbf{y'}\!-\!\mathbf{y_t^G}\!\Vert_2^2]_{+},
\label{margin}
\end{aligned}
\end{equation}
where $[\cdot]_+=max(0,\cdot)$, set $Y$ consists of the VD-oriented negative samples and $Y'$ consists of the stochastic negative samples.

\paragraph{Training for A-Bot.}
The training of A-Bot is the same as in Sec.~\ref{background_training}.


\subsubsection{Reinforcement Learning}
In RL, Q-Bot and A-Bot are jointly optimized with the reward assigned by AugG. 
At round $t$, AugG makes the image feature prediction $\mathbf{y^G_t}$. Then as Q-Bot questions $q_{t+1}$ and A-Bot answers $a_{t+1}$, AugG predicts $\mathbf{y^G_{t+1}}$.
Accordingly, Q-Bot and A-Bot are awarded by the reward:
\begin{gather}
\setlength{\abovedisplayskip}{3pt}
\setlength{\belowdisplayskip}{3pt}
r_t^G=\Vert \mathbf{y^{gt}} - \mathbf{y_{t}^G} \Vert_2^2 - \Vert \mathbf{y^{gt}} - \mathbf{y_{t+1}^G} \Vert_2^2.
\end{gather}

\section{Experiments}

We evaluate our method on the large-scale VisDial v1.0 dataset~\cite{Das2017VisualD}, where the train split contains 123,287 images and the validation split contains 2,064 images, and each image has the corresponding caption and 10-round dialog. 

For training details, please refer to Appendix~\ref{training_details}. The upper bound $B$ in the \textit{sampler} of ReeQ's Entity Selector is set to 1, and we discuss its effect and more options in Appendix~\ref{limit_sampling_rule}.


\subsection{Comparing Methods}
We compare our method with previous strong baselines. To clarify, we introduce the comparing methods in QGen and Guesser, w.r.t. the roles to generate questions or make image predictions.

\paragraph{QGen:} 1) DasQ~\cite{Das2017LearningCV}: the baseline method; 2) DivQ~\cite{Murahari2019ImprovingGV}: an improved method that penalizes the similarity of successive encoded dialog hidden states to encourage question diversity; 3) ReeQ: our method.

\paragraph{Guesser:} 1) DasG: Guesser that is cooperatively trained with DasQ; 2) DivG: Guesser that is cooperatively trained with DivQ; 3) AugG: our Augmented Guesser.

\begin{figure}[] 
\centering 
\includegraphics[width=.4\textwidth]{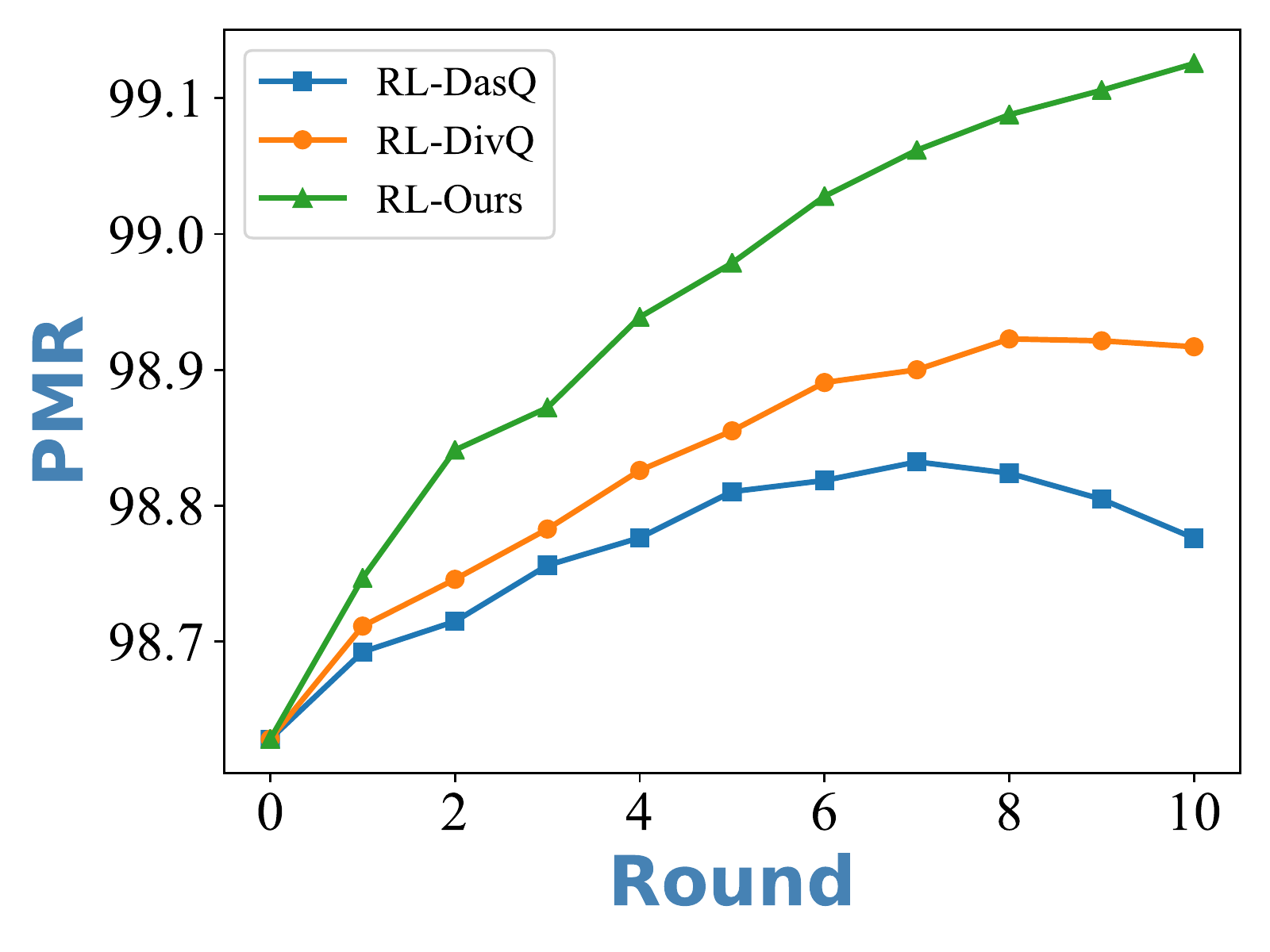} 
\caption{Trends of PMR as dialog progresses.}
\label{pmr}
\end{figure}

\subsection{Quantitative Results}

\paragraph{Image-Guessing Task.}
We evaluate the performance on image-guessing task. In concrete, QGen and A-Bot firstly generate 10-round dialog, then Guesser makes a prediction about the unseen image, lastly the candidate images (images in validation split) are sorted according to their similarity to the prediction and compute the rank of the target image. The evaluation metrics are: 1) MRR~\cite{2003Evaluating}: mean reciprocal rank of target image; 2) R@k~\cite{Das2017LearningCV}: the existence of target image in the top-k images; 3) Mean~\cite{Das2017LearningCV}: mean rank of target image; 4) PMR~\cite{Das2017LearningCV}: percentile mean rank. 

\begin{table*}[]
\centering
\setlength{\tabcolsep}{2.mm}{
\begin{tabular}{c|c|c|cccccc}
\hline
\#& & & Unique questions $\uparrow$& Mutual overlap $\downarrow$ & Dist-1 $\uparrow$& Dist-2 $\uparrow$ & Ent-1 $\uparrow$ & Ent-2 $\uparrow$\\ \hline
1&\multirow{3}{*}{SL} & DasQ$\ddagger$ & 6.57 &  0.60 & 2.70 & 3.00 & 0.34 &  0.42 \\
2&& DivQ$\ddagger$ & 7.45 & 0.51 & 2.82 & 3.18 & 0.38 & 0.48 \\
3&& ReeQ & \textbf{9.97} & \textbf{0.11} & 2.87 & 3.41  & \textbf{0.46}  & 0.63 \\ 
\hline
4&\multirow{3}{*}{RL} & DasQ$\ddagger$ & 6.70  & 0.58  & 2.72 & 3.03  & 0.35 & 0.43 \\
5&& DivQ$\ddagger$ & 8.19 & 0.41 & 2.90 & 3.31 & 0.40  & 0.53 \\
6&& ReeQ & \textbf{9.97} & \textbf{0.11} & \textbf{2.90} & \textbf{3.45} & \textbf{0.46} & \textbf{0.64} \\ \hline
\end{tabular}}
\caption{Question diversity on VisDial v1.0 val. $\ddagger$ means the results are cited from ~\cite{Murahari2019ImprovingGV}. $\uparrow$ indicates higher is better. $\downarrow$ indicates lower is better.}
\label{diversity}
\end{table*}

We illustrate the results in Tab.~\ref{image-guessing-performance}. As shown in row 10, our method achieves the best performance on all metrics and becomes the new state of the art, with a MRR of 33.65, R@1 of 19.91,R@5 of 48.50, R@10 of 62.94 and PMR of 99.13.
To make fair comparisons, we further use the same AugG as Guesser to evaluate all methods, as shown in row 8, 9 and 10. As can be seen, our method is superior than RL-DasQ and RL-DivQ on all metrics. 
As in SL (see row 3,4 and 5), SL-ReeQ outperforms other methods on R@5, R@10, Mean and PMR, but does not surpass SL-DivQ on MRR and R@1.
This may come from the accumulated errors of entity selection in the instances that are unseen in SL, as the training data only covers limited selecting trajectories. 
And since RL enables more explorations, the problem is relieved and RL-ReeQ achieves the best performance.

Fig.~\ref{pmr} shows the trends of PMR in the 10-round dialog. To make a fair comparison, we use the same AugG to serve as the Guesser. 
As can be seen, only our method enables the continuously increasing image-guessing performance as dialog progresses. The trends indicate that our method can generate the constantly visually-related and informative dialogs while others cannot.

\paragraph{Question Diversity.}
We evaluate the question diversity of Q-Bot with the following metrics: 1) Unique questions~\cite{Murahari2019ImprovingGV}: mean number of unique questions in the 10-round dialog; 2) Mutual overlap~\cite{DBLP:journals/corr/abs-1805-12589}: mean BLEU-4~\cite{2002A} overlap with the other 9 questions in the 10-round dialog; 3) Dist-n and Ent-n~\cite{2015A,2018Generating}: number and entropy of distinct n-grams in the generated questions divided by the total number of tokens. 

As shown in Tab.~\ref{diversity}, row 6 indicates that our method achieves the new SOTA performance on question diversity. Specifically, our RL-ReeQ achieves approximately 2 points improvement on Unique questions, which shows that we have greatly reduced repetition (row 4, 5 vs. row 6). Noticeably, our result on Unique questions is approaching the upper bound, i.e., 10. Besides, our method also achieves better language diversity according to Mutual overlap, Dist-1, Dist-2, Ent-1 and Ent-2 (row 3 and row 6).


\begin{table}[]
\small
\renewcommand\arraystretch{1.2}
\centering
\setlength{\tabcolsep}{0.4mm}{
\begin{tabular}{l|cccccc}
\hline
            & NDCG $\uparrow$ & MRR $\uparrow$ & R@1 $\uparrow$ & R@5 $\uparrow$ & R@10 $\uparrow$ & Mean $\downarrow$\\ \hline
SL$\ddagger$     & 53.10    & 46.21    & 36.11    & 55.82    & 62.22     & 19.58  \\
RL$\ddagger$    & 53.76    & 46.35    & 36.22     & 56.15   & 62.41     & \textbf{19.34}  \\ 
RL-Div$\ddagger$   & 53.91    & 46.46    & 36.31    & 56.26   & 62.53    & 19.35  \\
RL-ReeQ & \textbf{54.35}    & \textbf{46.52}    & \textbf{36.45}    & \textbf{56.34}    & \textbf{62.68}     & 19.56  \\
\hline
\end{tabular}}
\caption{A-Bot performance on VisDial v1.0 val. $\ddagger$ means the results are cited from ~\cite{Murahari2019ImprovingGV}. $\uparrow$: higher is better. $\downarrow$: lower is better.}
\label{A-BotResult}
\end{table}

\begin{table}[]
\small
\renewcommand\arraystretch{1.2}
\centering
\setlength{\tabcolsep}{0.7mm}{
\begin{tabular}{l|cccccc} \hline
& MRR $\uparrow$ & R@1 $\uparrow$ & R@5 $\uparrow$ & R@10 $\uparrow$ & Mean $\downarrow$ & PMR $\uparrow$ \\ \hline
DasG$\dagger$ & 3.29 & 11.58 & 20.06 & 9.17 & 108.11 & 94.76 \\
AugG$^-$ & 32.52 & 18.94 & 47.38 & 62.79 & 18.47 & 99.10 \\
AugG& \textbf{33.63} & \textbf{20.06} & \textbf{47.97} & \textbf{63.23} & \textbf{17.72} & \textbf{99.14} \\ \hline
\end{tabular}}
\caption{Guesser performance on VisDial v1.0 val. $\dagger$ represents that the evaluated model are from ~\cite{Murahari2019visdialdivpytorch}. AugG$^-$ means only stochastic negative samples are used in training.}
\label{image_guesser_performance}
\end{table}

\paragraph{A-Bot Performance.}
We evaluate the A-Bot performance in a retrieval setting, following ~\citet{Das2017VisualD}. Additional 100 candidate answers for each instance are provided and the model is evaluated by retrieval metrics: 1) NDCG~\cite{10.1145/582415.582418}: normalized discounted cumulative gain; 2) MRR~\cite{2003Evaluating}: mean reciprocal rank of the ground truth answer; 3) R@k~\cite{Das2017VisualD}: the existence of the ground truth answer in the top-k answers; 4) Mean~\cite{Das2017VisualD}: mean rank of the ground truth answer.
Tab.~\ref{A-BotResult} shows the comparing results of A-Bot performance. Our model achieves higher NDCG, MRR, R@1, R@5 and R@10.

\paragraph{Guesser Performance.}
Guesser performance is tested on the given ground-truth dialog, shown in Tab.~\ref{image_guesser_performance}. As can be seen, AugG achieves the best performance with a PMR of 99.14. By comparing AugG with AugG$^-$, we see that the performance is improved by VD-oriented negative samples.

\paragraph{Ablation Study.}
We conduct ablation study to investigate the effect of ReeQ and the effect of rewards given by different Guessers, respectively.
We use AugG as Guesser and evaluate the further image-guessing performance for fair comparisons.
As shown in Tab.~\ref{ablation_study}, we have following observations:
1) by comparing the results in upper part and lower part, we see the superiority of ReeQ;
2) in each part, by comparing the results among $+r_1$, $+r_2$ and $+r_3$, we see the respective improvements brought by separately optimized Guesser, as well as VD-oriented hard negatives in training Guesser. This indicates that our Guesser assigns more reliable rewards that help achieve improved performance in image-guessing.

\begin{table}[]
\small
\renewcommand\arraystretch{1.2}
\centering
\setlength{\tabcolsep}{0.6mm}{
\begin{tabular}{l|l|cccccc} \hline
\# & & MRR$\uparrow$ & R@1$\uparrow$ & R@5$\uparrow$ & R@10$\uparrow$ & Mean$\downarrow$ & PMR$\uparrow$ \\ \hline
1&DasQ+$r_1$  & 25.65 & 16.30 & 40.50 & 55.43 & 28.57 & 98.76 \\
2&DasQ+$r_2$ & 32.19 & 19.09 & 46.27 & 60.76 & 21.64 & 98.95 \\
3&DasQ+$r_3$ & 32.77 & 19.47 & 46.75 & 62.89 & 20.45 & 99.01 \\ \hline
4&ReeQ+$r_1$ & 32.27 & 18.56 & 46.75 & 61.01 & 19.58 & 99.05\\
5&ReeQ+$r_2$ & 32.78 & 19.38 & 47.00 & 62.65 & 19.46 & 99.06 \\
6& ReeQ+$r_3$ & \textbf{33.65} & \textbf{19.91} & \textbf{48.5} & \textbf{62.94} & \textbf{18.05} & \textbf{99.13} \\
\hline
\end{tabular}}
\caption{Performance of ablation methods on image-guessing task. $r_1$, $r_2$ and $r_3$ represent the reward is assigned by the cooperatively optimized guesser, AugG$^-$ and AugG, respectively.}
\label{ablation_study}
\end{table}

\begin{figure*}[]
\centering
\includegraphics[width=0.95\textwidth]{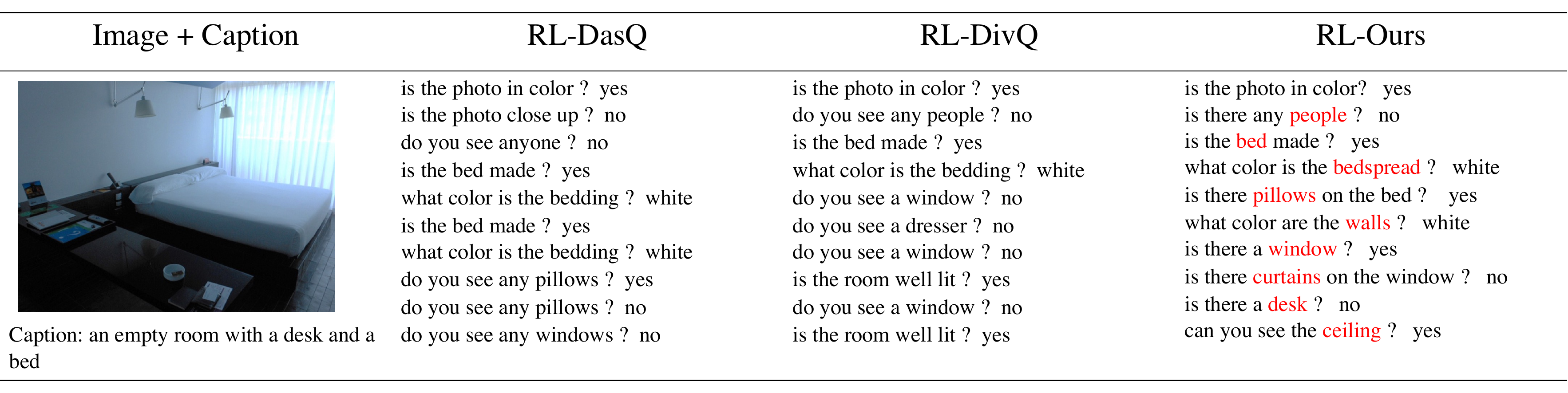}
\caption{An example of dialogs on VisDial v1.0 val. Red marks the entity selected at current round.}
\label{dialogIns}
\end{figure*}

\subsection{Qualitative Results}
Fig.~\ref{dialogIns} presents an example of generated dialogs from comparing methods.
As shown, both RL-DasQ and RL-DivQ ask repetitive questions while ours asks non-repetitive questions. Moreover, the questions generated by our method are more informative, detailed and of higher relevance to target image.
Noticeably, we find our method generates questions that are coherent with the selected entities (marked in red), indicating the entities guide the question generation effectively.
We also see the sequential entities follow a clear strategy. For example, it asks ``bed'' at round 3, then ``bedsprea'' and ``pillows on the bed''. Afterwards, it asks other furnishings in the room successively.
More qualitative results are given in Appendix~\ref{app_dialog_instance}.

\subsection{Human Study}
We conduct human studies to further evaluate the dialog generated by different methods, i.e, Human, RL-DasQ, RL-DivQ and RL-Ours. Six postgraduate students are recruited and each one evaluates 50 instances for each method.

\begin{table}[]
\small
\renewcommand\arraystretch{1.2}
\centering
\setlength{\tabcolsep}{1mm}{
\begin{tabular}{lcc} \hline
& R@1 $\uparrow$ & R@5 $\uparrow$ \\ \hline
Human & 75.00& 95.00 \\ \hline
RL-DasQ & 27.00 & 80.33 \\
RL-DivQ & 42.67 & 83.33 \\
RL-Ours & 46.33 & 89.00 \\
\hline
\end{tabular}}
\caption{Results on image-guessing in human study.}
\label{human_study_1}
\end{table}

\begin{figure} \centering    
\subfigure[Relevance] {
 \label{fig:a}     
\includegraphics[width=0.25\columnwidth]{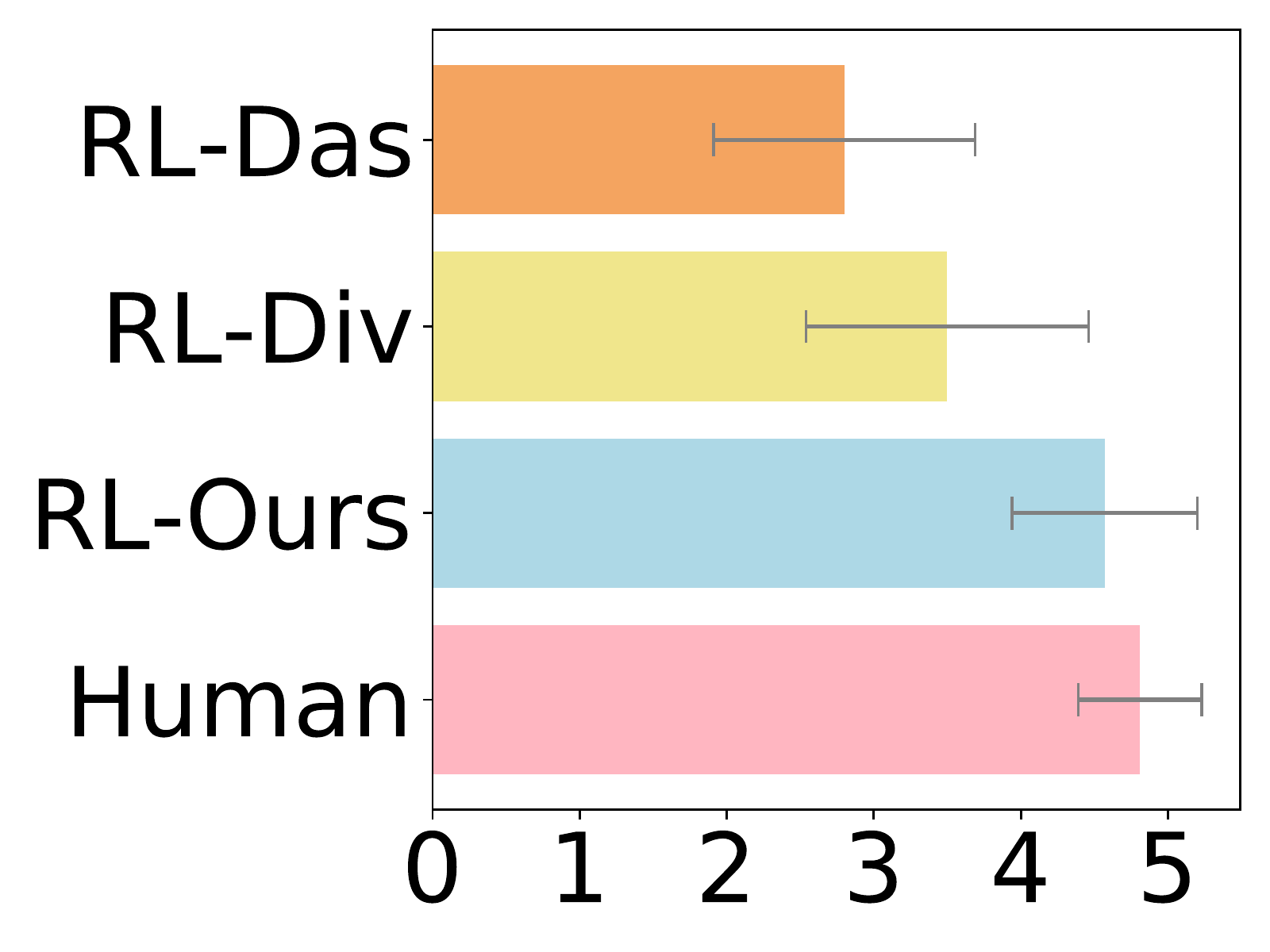}  
}     
\subfigure[Informativity] { 
\label{fig:b}     
\includegraphics[width=0.25\columnwidth]{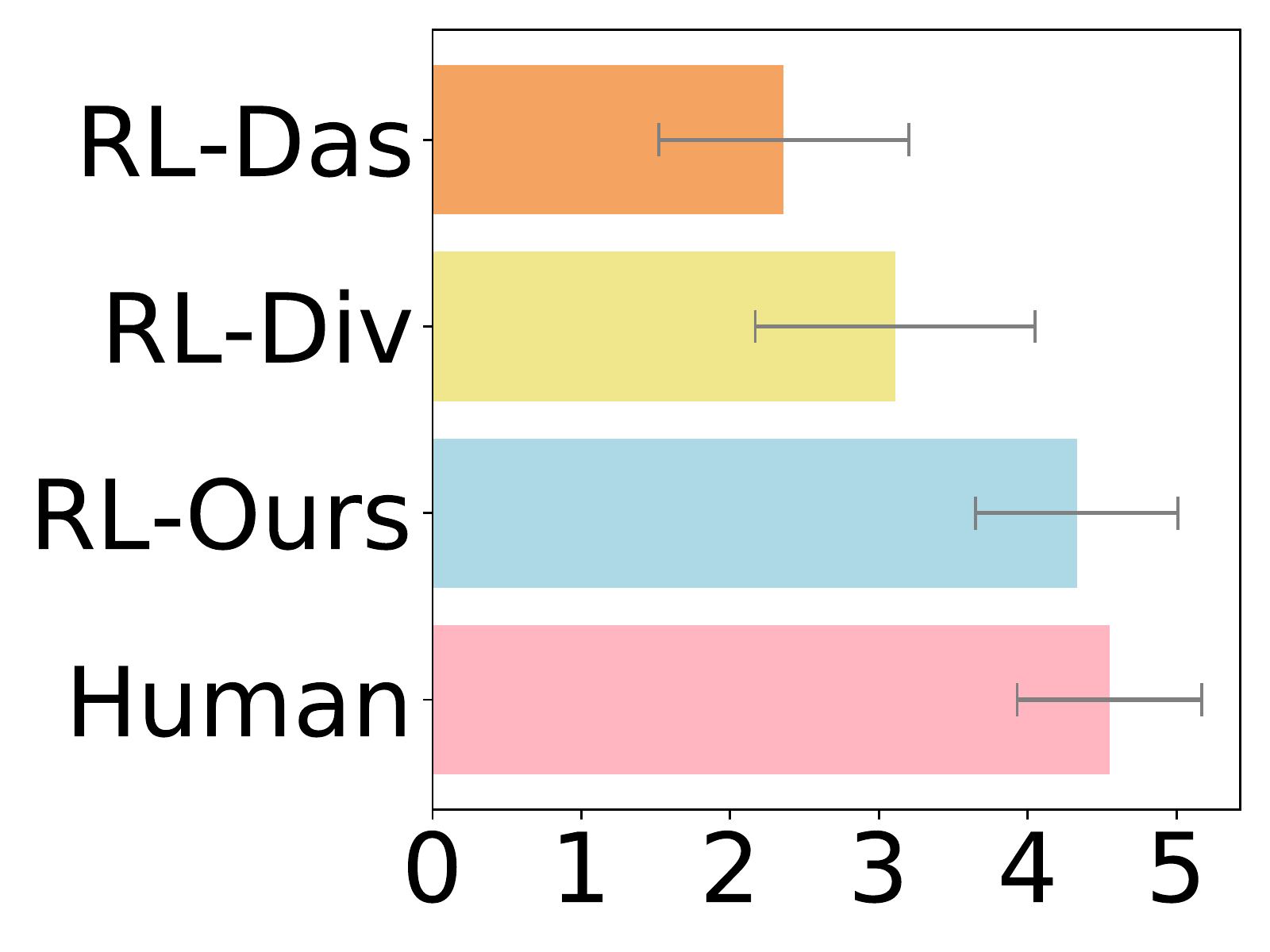}     
}    
\subfigure[Coherence] { 
\label{fig:c}     
\includegraphics[width=0.25\columnwidth]{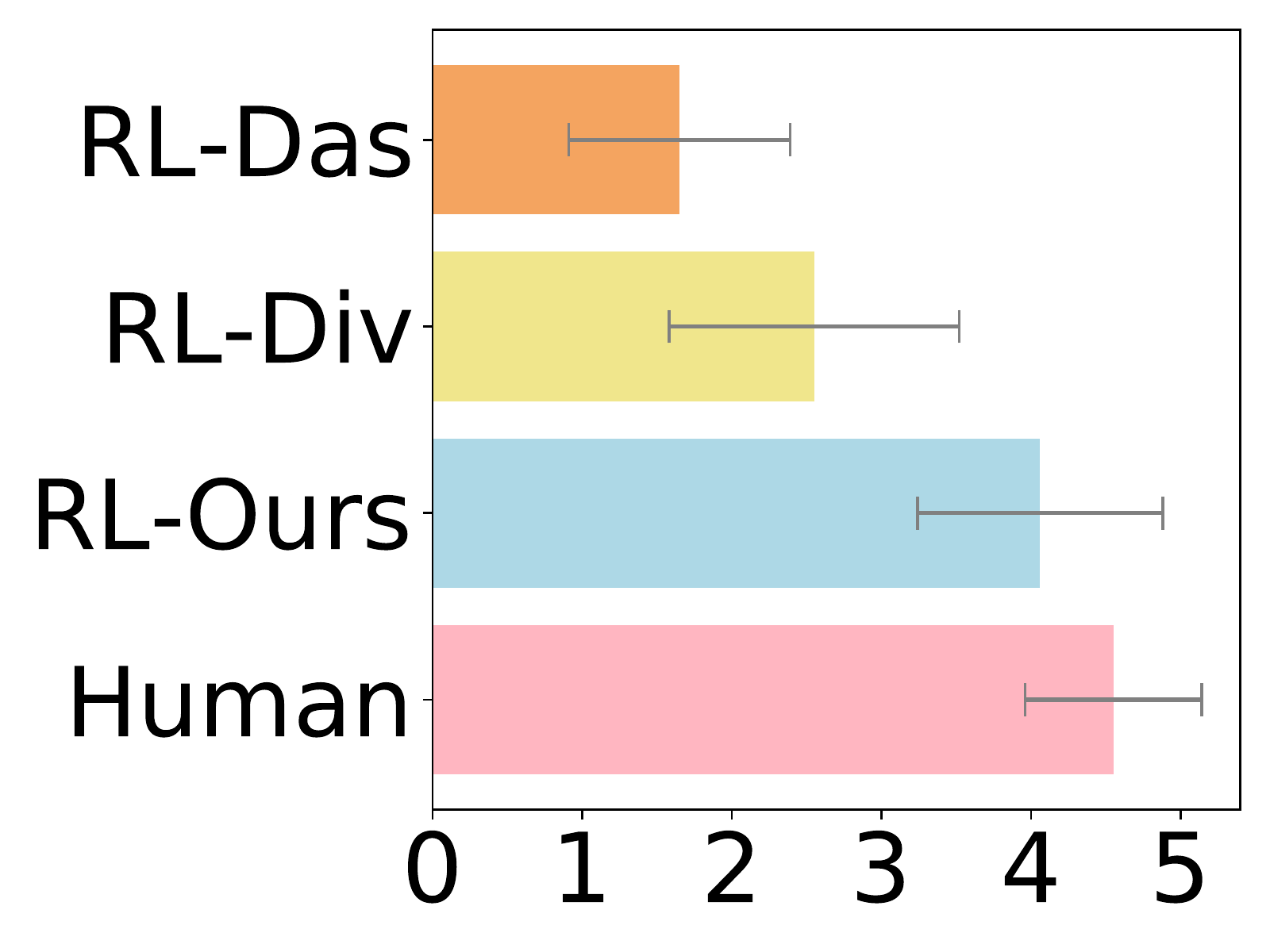}     
 }   
\caption{Results on dialog quality in human study, including means and variances.}     
\label{human_study_2} 
\end{figure}

\paragraph{Image-guessing Task.} \citet{Das2017LearningCV} design the task to evaluate how human-understandable and image-discriminative the generated dialogs are. Evaluators are required to pick the top-1 and top-5 likely images from a 16-candidate-image pool, including 1 target, 5 nearest neighbors and 10 random ones. As in Tab.~\ref{human_study_1}, our method outperforms previous work while has a gap with Human.

\paragraph{Dialog Quality.} Evaluators score the dialogs based on the dialog history and image. The scoring adopts a 5-point scale, which is evaluated in terms of relevance, informativity and coherence. Relevance indicates how well the generated dialogs are related to the target image and the caption. Informativity measures whether the dialog provides sufficient information related to the target image. Coherence assesses whether the generated dialogs are less repetitive, natural and coherent. As in Fig.~\ref{human_study_2}, humans judge our method as generating more visually related, informative and coherent dialogs than other methods.

\section{Discussion}
\paragraph{Entity-Selection Accuracy.} Considering the diverse questioning strategies in real-world scenes, "correct" entity is hard to define. Therefore, we conduct human study. We sample 50 generated dialogs (selected entities highlighted) in val-set and ask 3 evaluators to judge whether entities are relevant to the image and caption (i.e., are correct in the current context). 92.5\% selected entities are regarded as relevant (qualitatives in Appendix E).
Additionally, we study the ability to select entities just the same as entities in ground-truth human dialogs. Evaluation result on val-set shows 14\% selected entities are same, indicating the model has learned from human dialogs since there are 100 candidates.14\% is not high, but it is reasonable considering the rich visual scenes and various questioning paths.

\paragraph{Computational Cost.} Between ReeQ and DasQ, the ratio of time to get the best performed model is 1.47 (11h vs. 7.5h) in SL and 1.38 (14.5h vs. 10.5h) in RL. At inference, ReeQ spends 1.1 times the baseline (72s vs. 64s).
We conclude ReeQ costs more for estimating the entities to ask and generating entity-guided questions. Despite additional time cost, generation results of ReeQ are inspiring.

\section{Related Work}
Our work is mostly related to building open-domain Visual Dialog Questioner in the image-guessing task setting. \citet{Das2017LearningCV} propose the task and generate questions in a sequence-to-sequence fashion. \citet{Murahari2019ImprovingGV} propose to reduce repetition by penalizing the similarity in successive dialog hidden states. \citet{2019Building} retrieve the most-likely image, encode the image into a multi-modal context vector and use it to decode questions. These methods follow a sequence-to-sequence fashion while ReeQ explicitly uses related-entities as guidance to generate questions following a learned strategy.

Our work is also relevant to the works~\cite{DBLP:journals/corr/abs-1711-07614,8639546,2017End,2018Beyond,2019What,2020Answer} that focus on VD Questioner for GuessWhat?!~\cite{2016GuessWhat},  where the goal is to locate a target object in the image and the answers can only be ``yes/no/not available''. 
Compared to them, building a Questioner in a more open-domain VD setting is of more difficulty. Moreover, Q-Bot in GuessWhich has no access to visual information, making it harder to ask visually related questions.  

\section{Conclusion}
In this paper, we propose \textbf{R}elated \textbf{e}ntity \textbf{e}nhanced \textbf{Q}uestioner (ReeQ) and \textbf{Aug}mented \textbf{G}uesser (AugG) to enhance Visual Dialog Questioner in both SL and RL. ReeQ generates questions with related entities as guidance and learns an entity-based questioning strategy from human dialogs. AugG is a strong Guesser that is optimized for VD especially. We use AugG to assign reliable rewards in RL. Experimental results on VisDial v1.0 show our method outperforms priors on multiple criteria.

\section*{Acknowledgements}
We would like to thank anonymous reviewers for their suggestions and comments. The work was supported by the National Natural Science Foundation of China (NSFC62076032) and the Cooperation Poject with Beijing SanKuai Technology Co., Ltd.

\bibliographystyle{acl_natbib}
\bibliography{custom}

\clearpage
\appendix

\section{Related Entity Retrieval}\label{entity_retrieval}
This part we introduce related entity retrieval referred in Sec.~\ref{retriever}. As a prerequisite, We use the object dictionary in Visual Genome~\cite{2017Visual} as our entity vocabulary, and build entities-to-entities indexes from the entities in captions to the entities in dialogs. Then follows 4 steps:
\begin{enumerate}[leftmargin=*]
\setlength{\itemsep}{1pt}
\setlength{\parsep}{1pt}
\setlength{\parskip}{1pt}
\item[1)] Extract caption entities $E_c$ in the caption $c$ through template matching.
\item[2)] Retrieve probable entities $E_p$ by using entities in $E_c$ as queries from the established indexes.
\item[3)] Sort entities in $E_p$ according to the sum of co-occurrence frequency with the entities in $E_c$.
\item[4)] Retain the top-N entities to form a candidate entity set $E=\{e_1,e_2,\cdots, e_N\}$.

\end{enumerate}

As for details, We empirically set $N$ to 100. Averagely, 6.4 questions per 10-round dialog in the train split could match the candidate entities while 6.1 in the validation split. \textit{Selector} will choose an additional `NULL' when no entity could match with the question.

\section{VD-oriented Negative Samples}\label{negative_samples}
We obtain VD-oriented negative samples through the following steps.
Firstly, we build objects-to-images indexes through objects in the image, which are extracted using bottom-up-attention~\cite{Anderson_2018_CVPR}.
Secondly, we retrieve top-100 images through the index and the pre-trained model (we use OSCAR~\cite{2020Oscar}) successively.
Lastly, we sample 8 images from the retrieved images to form the VD-oriented negative samples, learning from prior work~\cite{2019Learning}.

\section{Training Details}\label{training_details}
We implement our method with Pytorch and conduct all experiments on NVIDIA Tesla V100 GPU.

Overall, we follow the same training methods with previous work. In SL, we pre-train ReeQ for 15 epochs. We use Adam optimizer with a mini-batch size of 20 and a learning rate of 1e-3 decayed to 5e-5. $\beta$ and $\gamma$ are set to be 1000 and 1. We also apply Dropout rate of 0.5 before the feature regression network as previous work.
For AugG, we train AugG for 10 epochs and select the best performed model on the validation set. Adam optimizer is used with a learning rate of 1e-3 and a batch size of 20. The margin $\alpha$ in $\mathcal{L}_{AugG}$ is set to 0.1 empirically.
And we directly use the released checkpoint of A-Bot from~\cite{Murahari2019visdialdivpytorch}.
In RL, we apply the same curriculum learning to fine-tune the model. Specifically, we use SL in the first K rounds of dialog, and optimize the model through RL in the remaining 10-K rounds. We start with K=9 and gradually decrease to K=4, and fine-tune the model for 12 epochs with a mini-batch size of 32. After each epoch, the model with the maximum PMR is selected for evaluation.

\section{Effect of Limit-sampling Rule}\label{limit_sampling_rule}
We investigate the effect of the limit-sampling rule.
In Tab.~\ref{limit_sampling_result}, when the upper bound $B$ (the maximum sampled times of each entity, as in Eq.~\ref{upb}) increases, the performance of image-guessing task and question diversity becomes lightly worse, validating the efficiency of limit-sampling rule in inference.
Besides, we study the absolute ability that ReeQ has learned through training with $w/ B=\infty$, which is equivalent to no limit-sampling rule is used in inference. 
As can be seen, ReeQ still achieves high unique questions of 8.87 and PMR of 99.05, indicating our ReeQ has acquired the ability to ask non-repetitive and visually related questions following the entity-based questioning strategy during training. 
And the limit-sampling rule further avoids the repetition through controlling the maximum sampled times of each entity, benefited from ReeQ following an entity-to-question fashion to generate questions.

\begin{table}[]
\small
\renewcommand\arraystretch{1.2}
\centering
\setlength{\tabcolsep}{0.5mm}{
\begin{tabular}{lcc|cc} \toprule
& \multicolumn{2}{c|}{Image-guessing} & \multicolumn{2}{c}{ Question diversity} \\ \hline
& PMR & MRR & Unique questions & Mutual overlap \\ \hline
$w/ B=1$ & 99.13 & 33.65  & 9.97 & 0.11\\
$w/ B=2$ & 99.06 & 33.50  & 9.18 & 0.25 \\
$w/ B=3$ & 99.06 & 33.31 & 9.03 & 0.26 \\ 
$w/ B=\infty$ & 99.05 & 33.03 & 8.87 & 0.28 \\
\bottomrule
\end{tabular}}
\caption{The effect of limit-sampling rule on image-guessing task. $w/ B=n$ means that the upper bound is set to be $n$ when ReeQ generates questions in inference. }
\label{limit_sampling_result}
\end{table}

\section{Qualitative Examples}\label{app_dialog_instance}
\begin{figure*}[h] 
\centering 
\includegraphics[width=1.\textwidth]{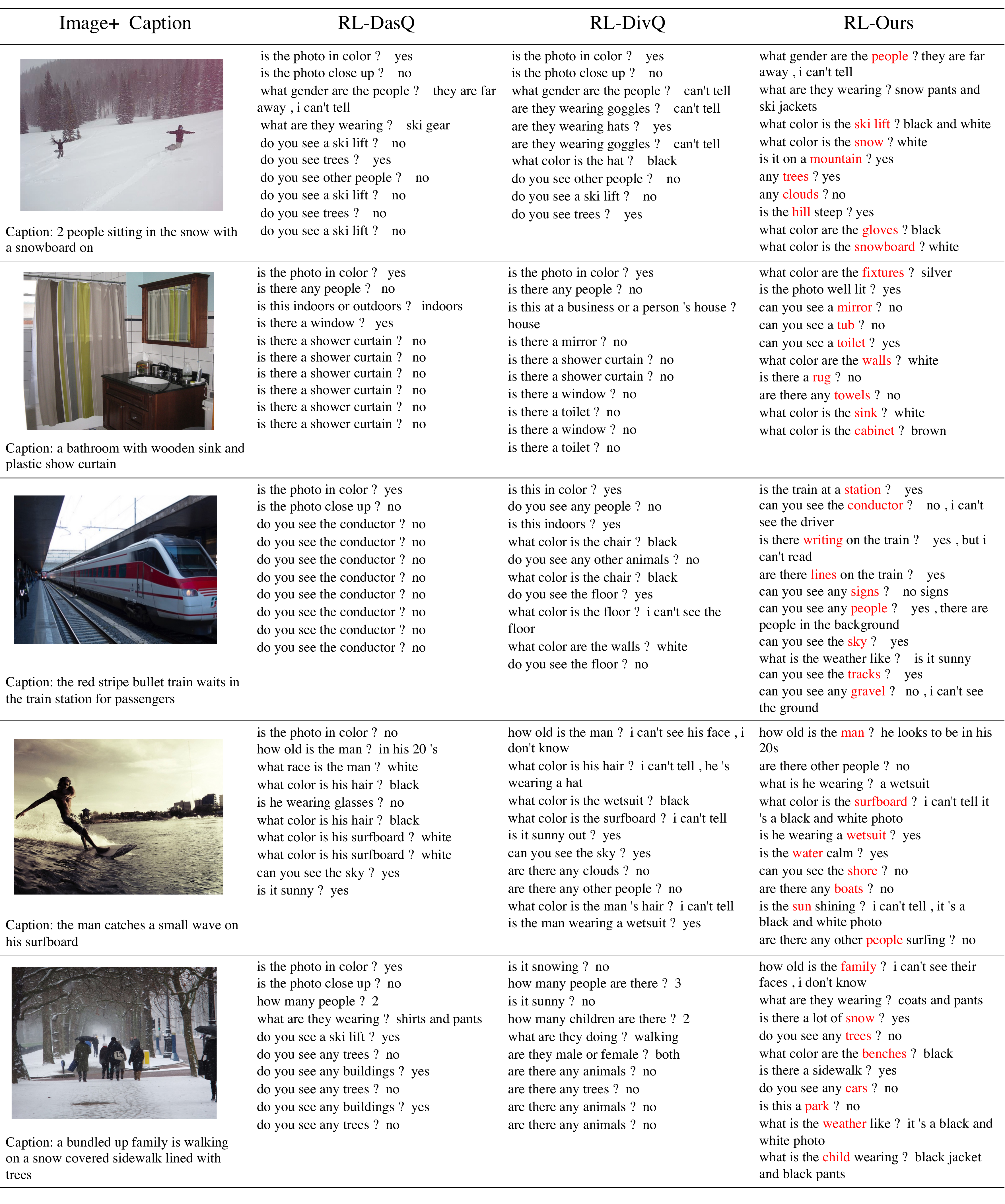} 
\caption{Selected examples of dialogs on VisDial v1.0 val. Red marks the entity selected at current round.} 
\label{dialogIns2} 
\end{figure*}

Fig.~\ref{dialogIns2} gives more examples of dialogs generated by different methods. RL-DasQ usually asks repetitive questions. RL-DivQ reduces repetition while generates less visually related questions. Comparatively, ours asks more informative and visually related questions. 
As marked in red, our method can select appropriate entity at each round and ask question accordingly.

\end{document}